\documentclass[letterpaper, 10 pt, conference]{ieeeconf}  

\IEEEoverridecommandlockouts                              

\usepackage{amsmath}
\usepackage{booktabs}
\usepackage{graphicx}   
\usepackage{subcaption} 
\let\labelindent\relax
\usepackage{enumitem}
\usepackage{cite}
\usepackage{amsfonts}
\usepackage{amssymb}
\usepackage{algorithm}
\usepackage{placeins}
\usepackage{marvosym}

\usepackage{amsthm}

\theoremstyle{definition}
\newtheorem{definition}{Definition}

\theoremstyle{plain}

\theoremstyle{remark}

\usepackage[table]{xcolor}
\usepackage{array}
\definecolor{oursblue}{RGB}{204,232,248}

\usepackage{soul}
\newcommand{\remove}[1]{}

\title{\LARGE \bf When to Probe and When to Act: Clarity-Aware Risk-Adaptive Control for Interactive Driving}

\author{
Youran Wang$^{1,*}$,
Devodita Chakravarty$^{2,*}$,
and Yiwei Lyu$^{3}$\textsuperscript{\Letter}\\[-2pt]
{\small $^{*}$Equal Contribution \qquad
\Letter~Corresponding Author}%
\thanks{$^{1}$Youran Wang is with the Departments of
Computer Sciences and Mathematics, University of
Wisconsin--Madison, USA.
{\tt\small wang3488@wisc.edu}}%
\thanks{$^{2}$Devodita Chakravarty is with the Department
of Mechanical Engineering, Indian Institute of
Technology Kharagpur, India.
{\tt\small devoditac@kgpian.iitkgp.ac.in}}%
\thanks{$^{3}$Yiwei Lyu is with the Department of Computer
Science and Engineering, Texas A\&M University, USA.
{\tt\small yiweilyu@tamu.edu}}%
}

\begin{document}

\IEEEaftertitletext{\vspace{-10pt}}
\maketitle
\thispagestyle{empty}
\pagestyle{empty}

\begin{abstract}

Safe navigation in interactive traffic requires autonomous vehicles to reason about unobservable human characteristics and the reciprocal nature of multi-agent interactions. Traditional architectures often decouple prediction from planning, leading to the “Frozen Robot Problem” or unsafe maneuvers under high ambiguity because they treat risk assessment and information acquisition as independent objectives.
We argue that the notion of safety should not remain static but must instead dynamically co-evolve with the system’s internal confidence: 
the robot should actively clarify ambiguity when necessary while also recognizing the diminishing returns of information, and naturally transition toward task execution.
We introduce a unified belief-space Model Predictive Control (MPC) framework that couples multistep information planning with a clarity-adaptive Conditional Value-at-Risk (CVaR) barrier mechanism. 
By treating clarity, a scalar informational state, as an explicit planning variable,
the controller co-designs its probing strategy with its safety specification, tightening conservatism under high ambiguity and relaxing constraints as confidence improves.
We further provide a performance analysis showing that the cost of information seeking and the expected planning-cost gap to a controller that knows the true human characteristic are bounded by residual ambiguity and solver error.
We evaluate the framework in lane-merging and unsignalized-intersection scenarios against representative interactive-planning baselines. Across 5,000 simulations, the proposed method achieves the highest success rates (94.6\% and 93.2\%) and the lowest collision rates (1.2\% and 1.4\%) among the compared methods, providing a principled foundation for robust decision-making in human-centric environments.

\end{abstract}

\vspace{-.2cm}
\section{Introduction}
\vspace{-.2cm}

Deploying autonomous vehicles into dense, unstructured traffic fundamentally changes the nature of the planning problem. Unlike structured environments where safety can be enforced through physical separation or geometric buffers~\cite{sadigh2018planning}, urban driving requires reasoning over the latent characteristics of surrounding human drivers, such as aggressiveness and attentiveness, none of which are directly observable and which vary across drivers~\cite{zhang2024intention}. This demands the ability to form, maintain, and act on probabilistic beliefs about hidden human goals in real time~\cite{zheng2022belief}.

This is most challenging in interactive scenarios such as lane merging and unsignalized intersections, where the ego vehicle's own motion shapes how surrounding drivers respond~\cite{chakravarty2026multimodal, cleac2019algames}. Classical ``predict-then-plan'' architectures sidestep this coupling by treating predicted human trajectories as fixed environmental inputs, routinely devolving into the \emph{Frozen Robot Problem}~\cite{trautman2010unfreezing, jasour2025risk} under persistent ambiguity. Fixed safety margins can further worsen the situation: even with improved prediction, substantial remaining belief uncertainty may keep the planner overly conservative, leading to deadlocks that prediction alone cannot resolve~\cite{zhang2023human}.

Covariance steering~\cite{zheng2022belief} and hierarchical belief-space models~\cite{chakravarty2026multimodal} improve the representation of interaction uncertainty by propagating joint ego-human beliefs and refining estimates of latent driver parameters through Bayesian inference. Yet when ambiguity persists due to sensor degradation, the planner must also actively select actions that generate informative observations~\cite{yin2025occlusion, sadigh2018planning, knaup2024active, hu2022active, gadginmath2025active}. A critical and underappreciated property of such information acquisition is that its marginal value diminishes as the belief sharpens, a structure commonly exploited through submodular objectives~\cite{rezazadeh2019sub, wang2020planning, zhang2025near}. A capable planner must therefore reason not only about \emph{how} to probe, but \emph{when to stop}. 
Prior information-seeking controllers~\cite{naveed2025provably, moon2026ia} have not resolved this transition endogenously, relying on explicit thresholds or continuing past the point of diminishing returns, an inefficiency that requires planning over a \emph{multistep} predictive horizon~\cite{naveed2025provably, moon2025informative} to correct.


Safety considerations introduce an orthogonal but inseparable challenge. Human-robot interaction exposes autonomous vehicles to rare, high-consequence events whose probability alone is an insufficient summary of risk~\cite{tangri2026value, kishida2024risk}. CVaR addresses this by quantifying expected cost in the worst-case tail, and its integration into predictive control yields tractable formulations that explicitly regulate extreme events~\cite{yin2023risk, lee2022risk}, with risk-adaptive~\cite{wang2025safe} and non-Gaussian~\cite{vahs2024risk} extensions. Yet a fundamental gap persists: existing risk-aware formulations generally specify safety margins independently of the vehicle's belief confidence. An agent under high ambiguity cannot reliably estimate its own risk, yet current frameworks assign it the same safety threshold as one with near-perfect knowledge, and conversely, overly conservative margins under high confidence can themselves trigger the frozen-robot failure mode.


The core insight motivating this work is that belief uncertainty should jointly determine the value of further information and the conservatism of safety regulation: persistent ambiguity motivates active probing, while diminishing information value reduces the incentive to probe as the belief sharpens. We introduce a belief-space MPC framework
in which a scalar informational state, \textit{clarity}, derived from the Riccati covariance recursion, serves as an endogenous planning variable coupling multistep information objectives to adaptive CVaR safety constraints within a single optimization.
The \textbf{main contributions} are:
\textbf{(1)} A belief-space MPC formulation treating clarity as an endogenous planning variable that couples uncertainty about the human characteristic to both the multistep information objective and safety constraints, with \textit{self-terminating probing} emerging naturally from the optimization as the diminishing-returns structure de-emphasizes probing as the belief sharpens.
\textbf{(2)} A co-design of probing strategy and safety notion through a clarity-adaptive CVaR barrier, wherein risk constraint conservatism is governed by the same belief state that drives information gathering, preventing simultaneous overconfidence in safety margins and underinformation about the human characteristic.
\textbf{(3)} A performance analysis showing that the cost of probing and the performance loss caused by uncertainty about the human characteristic are bounded by the remaining ambiguity and solver error.
\vspace{-.1cm}
\section{Proposed Methodology}
\label{sec:methodology}
\vspace{-.1cm}

In this paper, we aim to answer the following research questions: \textbf{Q1:} How should an autonomous vehicle represent and propagate uncertainty about the latent human characteristic so that planning can account for how its actions shape future beliefs?
\textbf{Q2:} When uncertainty persists during interaction, how can the ego vehicle plan informative actions and determine when to shift toward task execution as the value of further information diminishes?
\textbf{Q3:} How can safety be enforced under deep uncertainty by explicitly regulating tail risk in a manner that adapts to the vehicle's confidence?
To address these questions, we formulate a belief-space model predictive control framework that predicts how ego actions affect future belief covariance and its induced \emph{clarity}. By treating these as explicit planning variables over a finite horizon, the controller uses predicted clarity to jointly regulate the incentive for further probing and the conservatism of the CVaR safety constraint within a unified MPC optimization.\vspace{-.2cm}

\subsection{Belief-Space Modeling of Interactive Dynamics}
We consider a coupled ego--human system evolving in discrete time indexed by $k \in \mathbb{N}$ with uniform sampling period $\Delta t > 0$. The augmented state $x_k \in \mathbb{R}^{n_x}$ comprises the ego vehicle's physical states $(s_E,v_E)$, the human-driven agent's observable kinematic states, and a scalar latent parameter $\theta$ representing the human driver's characteristic. 
This latent parameter influences the coupled dynamics and captures uncertainty in the human driver's response.
The system dynamics are modeled via a locally linearized stochastic system:
$x_{k+1} = A_k x_k + B_k u_k + c_k + G_k w_k$,
where $u_k \in \mathbb{R}^{n_u}$ is the ego control input,
$A_k$ and $B_k$ are the corresponding dynamics Jacobians,
$c_k$ is the affine linearization term, and
$w_k \sim \mathcal{N}(0,I)$ represents process noise. This treatment retains nonlinear mean prediction while enabling tractable covariance propagation, as specified in \eqref{eq:belief_update}.

The observation model is given by $y_k = C x_k + D_k v_k$,
where $y_k \in \mathbb{R}^{n_y}$ denotes sensor measurements of the human-driven agent and $v_k \sim \mathcal{N}(0, I)$ is measurement noise; the ego's own state is known. The noise scale $D_k$ depends on relative geometry, while $\theta$
affects the human response dynamics; therefore, ego actions change both sensing quality and the informativeness of future observations about $\theta$, enabling active probing.

Under these assumptions, the belief over the system state remains Gaussian:
$b_k := \mathcal{N}(\hat{x}_k, P_k)$,
where $\hat{x}_k = \mathbb{E}[x_k \mid y_{0:k}]$ and $P_k \in \mathbb{S}_{+}^{n_x}$ is the estimation error covariance. The belief evolves according to the Kalman filter recursion:

\begin{equation}\footnotesize
\label{eq:belief_update}
\begin{aligned}
\hat{x}_{k|k-1} &= f(\hat{x}_{k-1}, u_{k-1}), \\
P_{k|k-1} &= A_{k-1} P_{k-1} A_{k-1}^\top + G_{k-1} G_{k-1}^\top, \\
P_k &= (I - L_k C) P_{k|k-1},
\end{aligned}
\end{equation}
with Kalman gain
\begin{equation}\footnotesize
L_k = P_{k|k-1} C^\top (C P_{k|k-1} C^\top + D_k D_k^\top)^{-1}.
\end{equation}

For a fixed initial belief and control sequence, the planner evaluates the model linearizations and noise scales along the nominal predicted trajectory. Conditioned on these quantities, the Riccati recursion produces a deterministic covariance prediction. This allows uncertainty evolution to be optimized alongside physical dynamics without introducing stochastic decision variables, thereby enabling tractable belief-space MPC.
\vspace{-.1cm}
\subsection{Clarity as an Informational State}
\vspace{-.1cm}
Under our Gaussian approximation, the covariance $P_k$ fully characterizes estimation uncertainty, but its matrix-valued nature complicates direct reasoning about confidence within an optimization loop.
Inspired by \cite{agrawal2023sensor}, we introduce a scalar informational state termed \emph{clarity}, defined over the interaction-relevant latent subspace as
\begin{equation}\footnotesize
\label{eq:clarity}
\beta_k \triangleq \left( 1 + \mathrm{tr}(P^{\theta}_k) \right)^{-1},
\end{equation}
where $P^{\theta}_k$ denotes the covariance block corresponding to the latent human characteristic $\theta$.
This transformation ensures $\beta_k \in (0,1]$, providing a bounded scale for jointly modulating information-seeking and safety conservatism. We define $\rho_k:=1-\beta_k$ as the residual ambiguity at step $k$,
representing the belief uncertainty that remains after incorporating
observations up to that step.

A main advantage of treating clarity as an explicit planning state instead of a post-hoc metric is that its evolution is fully induced by the predicted covariance recursion, ensuring consistency between estimation and information dynamics. Exposing $\beta_k$ to the planner increases the incentive for informative actions when clarity is low and shifts emphasis toward task execution as the value of further information diminishes.
The same $\beta_k$ also regulates safety conservatism: low clarity both increases the information-seeking incentive and tightens the clarity-adaptive CVaR constraint, whereas higher clarity reduces the value of further probing and relaxes the safety demand. Clarity therefore couples information acquisition and risk regulation within the same MPC optimization.
\vspace{-.2cm}
\subsection{Multistep Information Planning and Diminishing Returns}
Uncertainty in interactive driving cannot always be resolved instantaneously. When ambiguity persists, the ego vehicle must actively select actions that improve the quality of future observations. To this end, we quantify information acquisition through the reduction in belief uncertainty over a finite horizon.

Let $\mathcal{S} = \{u_k, \dots, u_{k+N-1}\}$ denote a candidate control sequence over horizon $N$. The expected information gain is defined as
\begin{equation}\footnotesize
\label{eq:info}
J_{\text{info}}(\mathcal{S}) \propto \mathrm{tr}(P^{\theta}_k) - \mathrm{tr}(P^{\theta}_{k+N}).
\end{equation}

This objective measures how much uncertainty in the latent human characteristic is reduced by executing $\mathcal{S}$. Because the latent parameter carries no process noise, the Riccati recursion makes $\mathrm{tr}(P^{\theta})$ non-increasing along a fixed candidate plan. To make the diminishing sensitivity explicit, write the scalar precision of the human characteristic as $\mathcal{I}=1/P^{\theta}$ and $\beta=\mathcal{I}/(1+\mathcal{I})$. For any increment $\Delta\mathcal{I}\ge0$,
\begin{equation}\footnotesize
\beta(\mathcal{I}+\Delta\mathcal{I})-\beta(\mathcal{I})
\le \Delta\mathcal{I}\,(1-\beta)^2,
\end{equation}
so the clarity gain produced by an equal precision increment decreases as the belief sharpens. This is the diminishing-returns property used below.

In our proposed MPC formulation, information acquisition is incorporated through the clarity variable, which serves as a smooth surrogate for multistep uncertainty reduction. Penalizing low clarity implicitly promotes actions that reduce future uncertainty, while naturally capturing diminishing returns as the belief converges without requiring explicit terminal information constraints.
This transition occurs automatically: as $\beta_k \to 1$, the clarity penalty 
$\Phi(\beta_k) = 1 - \beta_k$ in the MPC objective vanishes, causing the 
information-seeking term to contribute negligibly to the optimized cost. 
Consequently, the optimization places progressively less value on sacrificing task performance for further information as the belief sharpens, without requiring an explicit stopping threshold or mode-switching condition.
\vspace{-.2cm}
\subsection{Risk Modeling and Clarity-Adaptive CVaR Constraints}
\vspace{-.1cm}
While reducing uncertainty is important, interaction safety must be enforced even under worst-case outcomes. To capture tail risk explicitly, we model safety using Conditional Value-at-Risk (CVaR), which provides a coherent risk measure that directly regulates extreme events.

Let $Z_k = g(x_k) + \xi_k$
denote a stochastic safety margin, where $g(\cdot)$ is a signed safety metric (e.g., distance to collision boundaries) and $\xi_k$ is zero-mean Gaussian modeling error with variance $\sigma_\xi^2 > 0$. The safety margin $Z_k$ is defined such that smaller values indicate more severe safety violations, so that the \emph{lower-tail} CVaR at level $\alpha$ explicitly captures and penalizes worst-case (tail) risk outcomes. Under the Gaussian belief, $Z_k$ is Gaussian with mean $\mu_{Z_k} = g(\hat{x}_k)$ and variance $\sigma_{Z_k}^2 = \nabla g\, P_k \nabla g^\top + \sigma_\xi^2$, and the CVaR at tail probability $\alpha \in (0,1)$ admits the closed-form expression:
\begin{equation}\footnotesize
\label{eq:cvar}
\text{CVaR}_\alpha(Z_k) = \mu_{Z_k} - \sigma_{Z_k}
\frac{\phi(\Phi^{-1}(\alpha))}{\alpha}.
\end{equation}

To couple safety regulation with belief confidence, we introduce a 
clarity-adaptive CVaR barrier function defined over the 
stochastic safety margin $Z_k$.
\begin{definition}[Clarity-Adaptive CVaR Barrier Function]
The clarity-adaptive CVaR barrier function $h_k : \mathbb{R}^{n_x} 
\times \mathbb{S}^{n_x}_{+} \rightarrow \mathbb{R}$ is defined as
\begin{equation}\footnotesize
\label{eq:barrier}
    h_k = \mathrm{CVaR}_\alpha(Z_k \mid \hat{x}_k, P_k) 
    - \left( h_{\mathrm{safe}} + \kappa \left( \frac{1}{\beta_k} - 1 \right) \right),
\end{equation}
where $h_{\mathrm{safe}} > 0$ is a nominal safety threshold and 
$\kappa > 0$ is a confidence-weighting gain.
\end{definition}

This function satisfies the conditions for a valid discrete- time CBF:  it is continuously differentiable in $\hat{x}_k$, and the set
$\mathcal{C}_k=\{(\hat{x}_k,P_k)\mid h_k(\hat{x}_k,P_k)\geq0\}$
defines a belief-dependent safe region whose boundary is rendered forward invariant by the barrier condition introduced below. Here $Z_k$ is a signed safety margin, with smaller values indicating greater risk, and $\mathrm{CVaR}_\alpha$ denotes its lower-tail CVaR. Thus, $h_k\geq0$ implies $\mathrm{CVaR}_\alpha(Z_k)\geq h_{\mathrm{safe}}>0$, and hence $\mathrm{VaR}_\alpha(Z_k)>0$ and $\mathbb{P}(Z_k<0)\leq\alpha$ under the belief model.
The clarity-dependent term
$\kappa(1/\beta_k-1)=\kappa\,\mathrm{tr}(P^{\theta}_k)$
provides an additional ambiguity-dependent safety buffer: greater uncertainty about the human characteristic raises the required lower-tail CVaR margin, while the requirement approaches $h_{\mathrm{safe}}$ as $\beta_k\to1$.
Safety is enforced through the discrete-time barrier condition
\begin{equation}\footnotesize
\label{eq:barrier2}
h_{k+1} \ge (1 - \gamma(\beta_k))\, h_k - \zeta_k, \qquad \zeta_k \ge 0,
\end{equation}
where $\gamma(\cdot)\in(0,1)$ is a monotonically increasing Lipschitz function that modulates the allowable rate of risk relaxation and $\zeta_k$ is a slack variable priced linearly in the objective. Low clarity enforces slow relaxation to preserve safety under uncertainty, while increasing clarity permits a faster reduction in conservatism where $h_k\ge0$; when $h_k<0$ the same condition instead sets the rate at which the predicted margin must be recovered. This adaptive structure ensures a smooth transition from cautious probing to efficient execution while maintaining worst-case risk regulation throughout interaction.

\vspace{-.1cm}
\subsection{Unified Clarity-Risk-Aware MPC Formulation}\vspace{-.1cm}
The preceding components, belief-space dynamics, clarity-aware 
information planning, and risk-adaptive safety constraints, are 
unified into a single MPC formulation that enables joint reasoning 
over physical motion, information evolution, and worst-case risk 
over a finite horizon. At each decision step, the ego vehicle 
solves:\vspace{-.1cm}
\begin{equation}\footnotesize
\label{eq:unified_mpc}
\begin{aligned}
\min_{u_{0:N-1},\,\zeta_{0:N-1}} \ & \sum_{k=0}^{N-1}
\left( \ell_{\text{task}}(x_k, u_k) + \lambda_{\zeta} \zeta_k \right)
+ \frac{\lambda_{\text{info}}}{N}\sum_{k=0}^{N-1}\Phi(\beta_{k+1}) \\
& {}+ \lambda_{g} \left[ s_{\text{goal}} - s_{E,N} \right]_+^2 \\
\text{s.t. } \ &
x_{k+1} = f(x_k, u_k), \quad
P_{k+1} = \Phi_{\text{KF}}(P_k, x_k, u_k), \\
& \beta_{k} =
\left(1 + \mathrm{tr}(P^{\theta}_{k})\right)^{-1}, \\
& h_{k+1} \ge (1 - \gamma(\beta_k))\, h_k - \zeta_k, \quad \zeta_k \ge 0, \\
& u_{\min} \le u_k \le u_{\max}, \quad 
  0 \le v_{E,k} \le v_{\max},
\end{aligned}
\end{equation}
initialized at $x_0 = \hat{x}_k$, $P_0 = P_k$, with 
$\ell_{\text{task}}(x,u) = w_v (v_E - v_{\text{des}})^2 + w_u u^2$.
The objective balances four competing terms. The tracking term 
promotes task efficiency. The clarity penalty $\tfrac{\lambda_{\text{info}}}{N}\sum_{k}\Phi(\beta_{k+1})$, with $\Phi(\beta)=1-\beta$, acts as a surrogate for multistep information gain: it incentivizes 
uncertainty reduction when confidence is low and vanishes 
naturally as $\beta_k \rightarrow 1$, so the information-seeking incentive vanishes in the high-clarity limit without explicit mode logic. The slack variable 
$\lambda_\zeta$ keeps the barrier condition soft, and the goal term 
pulls the plan toward the goal by the end of the horizon.
Together, the formulation makes belief confidence an endogenous planning quantity rather than an externally monitored signal.
The resulting optimization is a structured nonlinear program with deterministic covariance propagation and closed-form risk terms that can be solved using standard nonlinear MPC solvers.

\vspace{-.2cm}
\subsection{Performance Analysis}

This analysis addresses two questions: \textbf{(1)} how much task-and-safety performance the controller may trade for information, and \textbf{(2)} how uncertainty about the human characteristic affects performance relative to knowing its true value. Let $p_k=(\hat{x}_k,P_k)$ denote the current belief, $U$ a candidate control sequence, and $J_k(U;p_k)$ the MPC cost excluding the information term.

\noindent\textbf{Assumptions.}
We assume: \textbf{(A1)} the human characteristic $\theta$ is fixed over the interaction, and its predicted variance is non-increasing, i.e., $P_{k+j}^\theta(U)\le P_k^\theta$; \textbf{(A2)} the posterior variance is calibrated, such that $\mathbb E[(\hat{\theta}_k-\theta)^2\mid\mathcal F_k]\le P_k^\theta$; and \textbf{(A3)} the applied solution is $\epsilon_k$-optimal, where $\epsilon_k\ge0$ represents solver error.
For \textbf{(A4)}, the proposed and reference controllers share the same compact control domain and hard constraints. Belief-dependent safety constraints use nonnegative, unbounded slack, so every $U$ can be evaluated under either belief. Let $p_k^{\mathrm{ref}}$ denote the belief obtained by replacing $\hat{\theta}_k$ with $\theta$ and setting its variance and cross-covariances to zero. We assume
\begin{equation}\footnotesize
\begin{gathered}
\left|J_k(U;p_k)-J_k(U;p_k^{\mathrm{ref}})\right|\le d_k,\\
d_k:=L_1\left|\hat{\theta}_k-\theta\right|
+L_2\left(P_k^{\theta}
+\left\|P_{-\theta,\theta,k}\right\|_2\right).
\end{gathered}
\label{eq:belief_cost_bound}
\end{equation}
These assumptions reflect the setting considered here: the human characteristic is locally stable, the filter covariance represents its estimation accuracy, the solver error is bounded, and the finite-horizon cost varies smoothly over the bounded planning region.

\vspace{.1cm}
We first argue that the cost of seeking information is limited by the ambiguity that remains. Define the predicted ambiguity cost as
$\Psi_k(U;p_k):=\frac{1}{N}\sum_{j=1}^{N}(1-\beta_{k+j}(U))$ and the complete objective as
$F_k(U;p_k):=J_k(U;p_k)+\lambda_{\mathrm{info}}\Psi_k(U;p_k)$, with $\lambda_{\mathrm{info}}\ge0$.
Let $U_k^\star$ be the $\epsilon_k$-optimal solution of $F_k$, and let
$U_k^0\in\arg\min_U J_k(U;p_k)$ be the solution obtained without the information term. Finally, let $\rho_k:=1-\beta_k$ denote the current residual ambiguity.

\vspace{.1cm}
\noindent\textbf{Proposition 1 (Information-seeking cost).}
Under (A1)--(A4),
\begin{equation}\footnotesize
0\le
J_k(U_k^\star;p_k)-J_k(U_k^0;p_k)
\le\lambda_{\mathrm{info}}\rho_k+\epsilon_k.
\end{equation}

\noindent\emph{Proof.}
Since $U_k^0$ minimizes $J_k$, the cost difference is nonnegative. The $\epsilon_k$-optimality of $U_k^\star$ bounds it above by
$\lambda_{\mathrm{info}}[\Psi_k(U_k^0;p_k)-\Psi_k(U_k^\star;p_k)]+\epsilon_k$.
Under (A1), predicted clarity cannot fall below its current value, so
$0\le\Psi_k(U;p_k)\le1-\beta_k=\rho_k$, which proves the result.
\hfill$\square$

The proposition has a direct interpretation: when the human characteristic is uncertain, the controller may accept a larger short-term cost to obtain useful observations. As clarity improves, the maximum cost attributable to information seeking decreases. The controller therefore shifts naturally from learning toward task execution.

Proposition~1 shows that our proposed controller does not incur excessive cost for probing. However, a small probing cost alone does not ensure good performance, since the controller still plans using an uncertain estimate of the human characteristic. We therefore compare it with the same planner given the true characteristic. This comparison connects belief quality to decision quality: calibration and cost smoothness bound the effect of estimation error, while Proposition~1 bounds the cost of probing. Together, they yield the following bound on the total performance loss caused by incomplete knowledge of the driver.
Let $U_k^{\mathrm{ref}}\in\arg\min_U J_k(U;p_k^{\mathrm{ref}})$ and define
$\Delta_k:=J_k(U_k^\star;p_k^{\mathrm{ref}})
-J_k(U_k^{\mathrm{ref}};p_k^{\mathrm{ref}})\ge0$.
Over $T$ replanning steps, let
$G_T:=\sum_{k=0}^{T-1}\Delta_k$ and
$A_T:=\mathbb E[\sum_{k=0}^{T-1}\rho_k]$.
\vspace{.1cm}

\vspace{.1cm}
\noindent\textbf{Theorem 1 (Residual-ambiguity cost bound).}
Suppose (A1)--(A4) hold, $\beta_k\ge\beta_{\min}>0$, and
$\sqrt{\operatorname{tr}(P_k)}\le\bar c$. Then
\begin{equation}\footnotesize
\mathbb E[G_T]
\le
C_1A_T+C_2\sqrt{T A_T}
+\mathbb E\!\left[\sum_{k=0}^{T-1}\epsilon_k\right],
\end{equation}
where
\begin{equation}\footnotesize
C_1=
\lambda_{\mathrm{info}}
+\frac{2L_2}{\beta_{\min}},
\qquad
C_2=
\frac{2(L_1+L_2\bar{c})}{\sqrt{\beta_{\min}}}.
\end{equation}

\noindent\emph{Proof.}
Adding and subtracting the costs evaluated under $p_k$, then applying Proposition~1 and (A4), gives
$\Delta_k\le\lambda_{\mathrm{info}}\rho_k+\epsilon_k+2d_k$.
Positive semidefiniteness of $P_k$ implies
$\|P_{-\theta,\theta,k}\|_2\le\bar c\sqrt{P_k^\theta}$.
Moreover, $P_k^\theta=\rho_k/\beta_k\le\rho_k/\beta_{\min}$, and (A2) gives
$\mathbb E[|\hat{\theta}_k-\theta|\mid\mathcal F_k]
\le\sqrt{\rho_k/\beta_{\min}}$.
Combining these relations yields
\begin{equation}\footnotesize
\mathbb E[\Delta_k\mid\mathcal F_k]
\le C_1\rho_k+C_2\sqrt{\rho_k}+\epsilon_k.
\end{equation}
Summing over $k$ and applying
$\sum_k\sqrt{\rho_k}\le\sqrt{T\sum_k\rho_k}$ and Jensen's inequality proves the result.
\hfill$\square$

The theorem shows that the performance gap is governed by accumulated ambiguity and solver error. The term $C_1A_T$ captures effects that scale directly with ambiguity, including information seeking, while $C_2\sqrt{T A_T}$ captures the effect of estimation error. If $A_T=o(T)$ and
$\mathbb E[\sum_{k=0}^{T-1}\epsilon_k]=o(T)$, then
\begin{equation}\footnotesize
\frac{\mathbb E[G_T]}{T}\longrightarrow0.
\end{equation}
Together, Proposition~1 and Theorem~1 show that, under the proposed probe-and-act strategy, the performance loss caused by uncertainty about the human characteristic is bounded by the remaining ambiguity and solver error. As observations improve clarity, both the cost of probing and the gap to a controller that knows the human characteristic decrease. This establishes a principled transition from information gathering to efficient task execution without an explicit switching rule.
\begin{figure*}[t]
    \centering
    \includegraphics[width=1\linewidth]{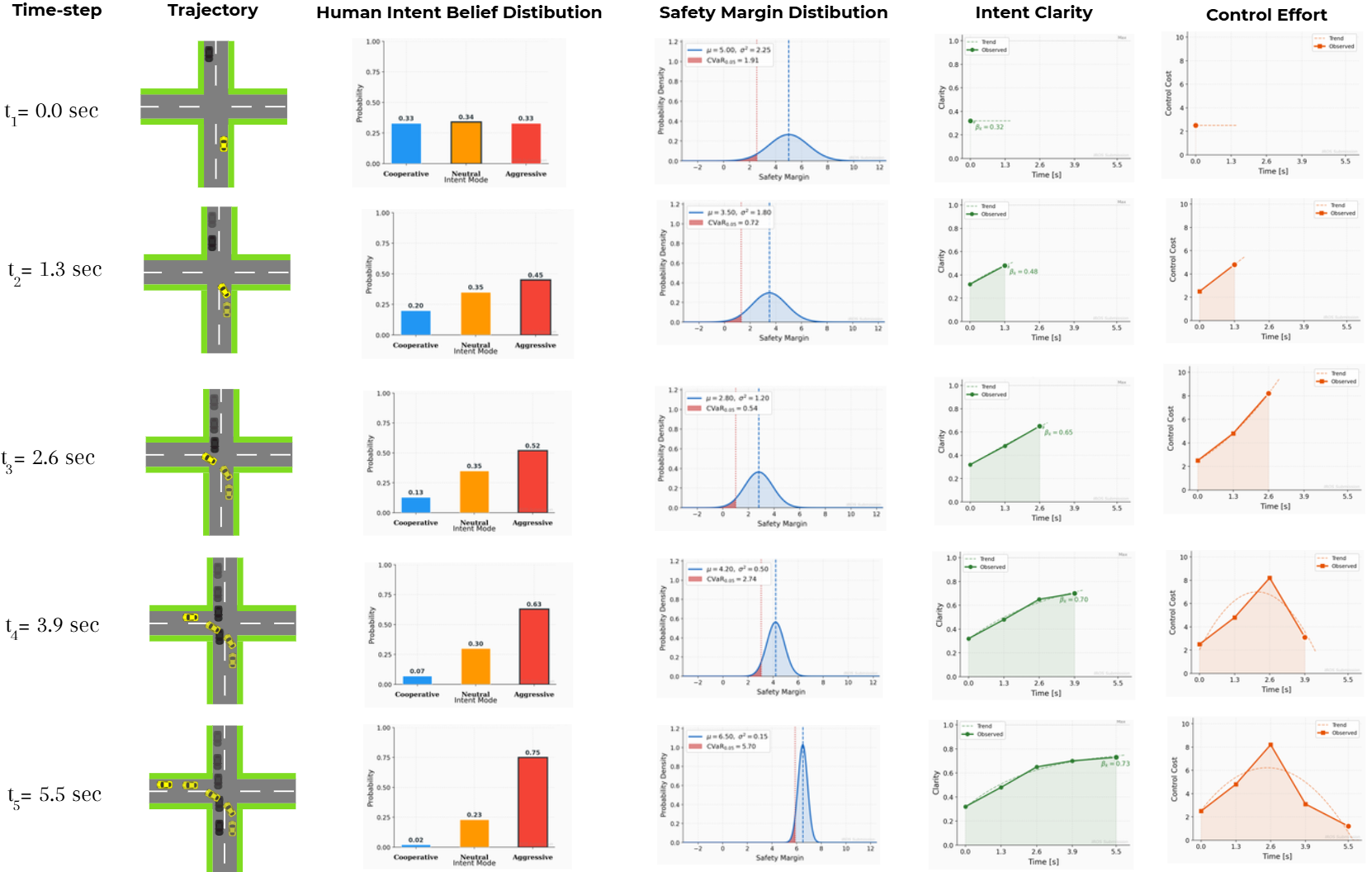}
    \caption{\footnotesize
\textbf{Timestep-by-timestep overview of the proposed framework executing an unsignalized intersection crossing.}
Yellow and black denote the ego and human-driven vehicles, respectively; translucent silhouettes show past positions. Each row presents vehicle trajectories, the belief over the human characteristic $\theta$ summarized over cooperative, neutral, and aggressive regions, the safety-margin distribution with lower-tail $\mathrm{CVaR}_{0.05}$, clarity $\beta_k$, and control effort.
At $t_1=0.0$~s, the ego approaches the intersection with a nearly uniform belief over $\theta$ and a broad safety-margin distribution. Low clarity yields a large safety demand
$h_{\text{safe}}+\kappa(1/\beta_k-1)$
in Eq.~\eqref{eq:barrier}.
At $t_2=1.3$~s, a cautious forward probe is followed by a shift toward an aggressive characteristic. The safety-margin distribution narrows and shifts left as the vehicles approach the conflict region, while clarity and control effort increase.
At $t_3=2.6$~s, the belief further favors an aggressive characteristic. Increasing clarity reduces both the additional safety buffer and the information penalty $\Phi(\beta_k)=1-\beta_k$, supporting a shift toward task execution.
At $t_4=3.9$~s, the belief over $\theta$ is more concentrated and the safety-margin distribution becomes sharper as the crossing progresses.
At $t_5=5.5$~s, the ego has cleared the intersection, with a narrow predicted safety-margin distribution and reduced control effort.
The sequence illustrates how evolving belief confidence jointly shapes information-seeking and safety conservatism, supporting a transition from cautious probing to task execution without explicit mode switching.}
    \label{fig:placeholder}
\end{figure*} \vspace{-.2cm}

\section{Simulation \& Discussion}
\label{sec:results}
\vspace{-.1cm}
\subsection{Baseline Comparisons}
\vspace{-.1cm}

We compare the proposed framework with four baselines spanning
risk-aware and information-seeking control.
\textbf{RA-MPPI }\cite{yin2023risk} is a sampling-based MPC method that incorporates CVaR for tail-risk regulation but treats the environment as reactive, operating under fixed uncertainty assumptions without modeling the latent human characteristic or actively reducing ambiguity. \textbf{AIG-MPC }\cite{sadigh2018planning} explicitly models the effect of robot actions on human responses and probes for informative observations, but enforces fixed safety constraints independent of belief confidence, leaving information acquisition and risk regulation loosely coupled.
\textbf{D-MPPI}~\cite{knaup2024active} combines
path-integral control with Bayesian inference to account
for the dual effect of actions on vehicle motion and
learning other drivers' behavioral parameters.
\textbf{ID-SMPC}~\cite{hu2022active}
uses shielding-aware dual control to reduce predictive
uncertainty while accounting for potential safety-filter
interventions.

\textbf{Implementation and tuning.}
The MPC problem is solved with IPOPT via CasADi. All baselines are reimplemented and adapted to the shared driving
benchmark. ID-SMPC additionally follows the authors' codebase,
with modifications to the human-response model.
All methods receive the same ego-state information and noisy
observations of the other vehicle, and share the vehicle dynamics,
actuator limits, and evaluation criteria.
Hyperparameters for RA-MPPI, AIG-MPC, and ID-SMPC are tuned
on separate validation runs, with configurations selected based
on success rate, safety, and completion time. D-MPPI uses the
parameter values reported in the original paper.

The baseline comparison experiments are conducted in two representative interactive driving scenarios: lane merging and unsignalized intersections. These scenarios capture distinct interaction structures, longitudinal negotiation under speed mismatch and lateral conflict resolution with right-of-way ambiguity, allowing us to evaluate whether performance trends remain consistent across different interaction geometries. Initial vehicle positions are randomized around the merge point or intersection center to avoid bias toward favorable geometries. Scenario-specific velocity ranges are used to induce diverse interactions: the ego vehicle travels at 7-10~m/s with surrounding traffic at 8-12~m/s in lane merging, while in intersections the ego velocity is 4-6~m/s and the crossing agent travels at 5-8~m/s. Both scenarios use $\Delta t = 0.1\,\text{s}$, a CVaR level $\alpha = 0.05$ and 500 runs per method; horizon lengths are $N = 40$ (merge) and $N = 60$ (intersection). 

\begin{figure*}
    \centering
\includegraphics[width=0.85\linewidth]{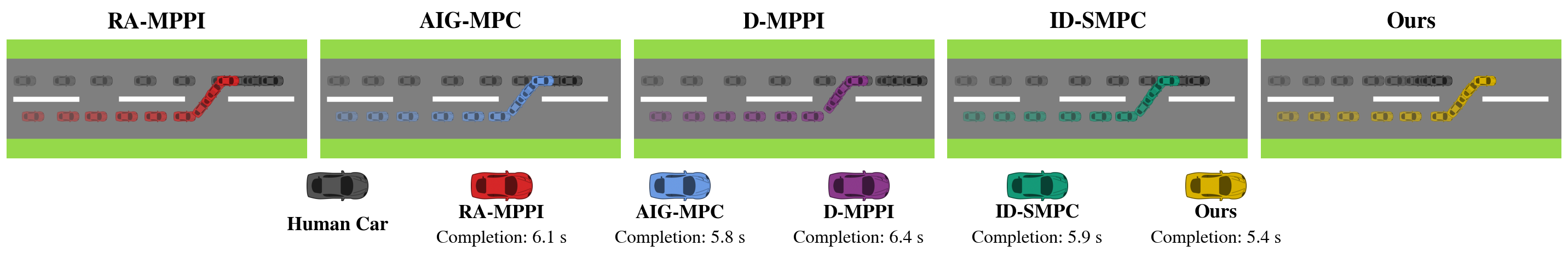}
    \caption{\footnotesize Qualitative lane-merging comparison in a selected episode. Faded vehicles indicate historical poses, while solid vehicles show the states at lane-change completion. The proposed method is the only method that merges ahead of the cooperative human driver in this episode, achieving the shortest task completion time of 5.4~s. RA-MPPI, AIG-MPC, D-MPPI, and ID-SMPC merge behind the human, completing the task in 6.1~s, 5.8~s, 6.4~s, and 5.9~s, respectively. }
    \label{fig:placeholder}
\end{figure*}

\begin{table*}[t]
\centering
\caption{\footnotesize
Quantitative comparison across lane-merge and unsignalized-intersection scenarios (500 runs per method per scenario).
The proposed method achieves the highest success rates (94.6\% and 93.2\%) and lowest collision rates (1.2\% and 1.4\%), respectively, together with the highest mean clarity and information gain in both scenarios.
D-MPPI has the shortest completion times, while AIG-MPC uses the least control effort.
RA-MPPI achieves the largest CVaR margin in lane merging, whereas the proposed method achieves the largest margin at intersections.
}
\label{tab:baseline_comparison}

\setlength{\tabcolsep}{2.6pt}
\resizebox{\textwidth}{!}{
\begin{tabular}{lccccccccc}
\toprule
\textbf{Method}
& \textbf{Success Rate$\uparrow$}
& \textbf{Collision Rate$\downarrow$}
& \textbf{Average Time (s) $\downarrow$}
& \textbf{Clarity $\uparrow$}
& \textbf{Worst-tail CVaR$_{0.05}$ Margin (m) $\uparrow$}
& \textbf{Information Gain (nats) $\uparrow$}
& \textbf{Control Effort $\downarrow$}
& \textbf{Solver Time (s) $\downarrow$} \\
\midrule

\multicolumn{9}{c}{\textit{Lane Merge}} \\
\midrule

\rowcolor{oursblue}
\textbf{Ours}
& $\mathbf{94.6\%}$
& $\mathbf{1.2\%}$
& $5.968 \pm 0.463$
& $\mathbf{0.841 \pm 0.082}$
& $1.509$
& $\mathbf{0.581 \pm 0.313}$
& $1.000 \pm 0.468$
& $0.32$ \\

D-MPPI
& $92.6\%$
& $1.4\%$
& $\mathbf{5.500 \pm 0.731}$
& $0.690 \pm 0.103$
& $0.813$
& $0.190 \pm 0.352$
& $0.690 \pm 0.119$
& $\mathbf{0.08}$ \\

AIG-MPC
& $86.6\%$
& $3.0\%$
& $5.718 \pm 0.182$
& $0.616 \pm 0.000$
& $-3.024$
& $0.188\pm 0.230$
& $\mathbf{0.094 \pm 0.027}$
& $1.24$ \\

RA-MPPI
& $85.6\%$
& $3.2\%$
& $6.059 \pm 0.502$
& $0.650 \pm 0.057$
& $\mathbf{2.092}$
& $0.078 \pm 0.193$
& $1.054 \pm 0.210$
& $0.54$ \\

ID-SMPC
& $91.4\%$ 
& $1.6\%$
& $5.726 \pm 0.194$
& $0.616 \pm 0.0178$
& $-2.518$
& $0.122 \pm 0.252$
& $0.810 \pm 0.686$
& $0.82$ \\

\midrule
\multicolumn{9}{c}{\textit{Unsignalized Intersection}} \\
\midrule

\rowcolor{oursblue}
\textbf{Ours}
& $\mathbf{93.2\%}$
& $\mathbf{1.4\%}$
& $7.452 \pm 0.603$
& $\mathbf{0.828 \pm 0.096}$
& $\mathbf{0.883}$
& $\mathbf{0.555 \pm 0.335}$
& $1.049 \pm 0.509$
& $0.65$ \\

D-MPPI
& $91.6\%$ 
& $1.6\%$
& $\mathbf{6.475 \pm 0.687}$
& $0.707 \pm 0.110$
& $-0.354$
& $0.242 \pm 0.398$
& $0.682 \pm 0.116$
& $\mathbf{0.26}$ \\

AIG-MPC
& $84.4\%$
& $3.2\%$ 
& $7.140 \pm 0.173$
& $0.616 \pm 0.000$
& $-2.313$
& $0.150 \pm 0.177$
& $\mathbf{0.152 \pm 0.051}$
& $1.35$ \\

RA-MPPI
& $82.6\%$ 
& $3.8\%$
& $7.800 \pm 0.570$
& $0.640 \pm 0.063$
& $0.405$
& $0.061 \pm 0.182$
& $1.070 \pm 0.363$
& $0.49$ \\

ID-SMPC
& $90.4\%$ 
& $1.6\%$
& $8.648 \pm 1.334$
& $0.642 \pm 0.078$
& $-0.342$
& $0.106 \pm 0.339$
& $1.321 \pm 0.942$
& $2.12$ \\

\bottomrule
\end{tabular}}
\end{table*}

\textbf{Table~I} summarizes quantitative results for lane merging and intersection scenarios. In the lane merge scenario, the proposed method achieves the highest success rate (94.6\%) and lowest collision rate (1.2\%), together with the highest mean clarity and information gain. D-MPPI achieves the shortest average completion time (5.50~s vs.\ 5.97~s for Ours), but has a lower success rate and slightly higher collision rate. AIG-MPC requires the least control effort, although its lower success rate and negative worst-tail CVaR margin indicate limitations in safety performance. RA-MPPI achieves the largest worst-tail CVaR margin, but exhibits a higher collision rate and lower success rate, indicating that this margin alone does not fully characterize task-level safety. Similar trends appear in the intersection scenario: the proposed method achieves the highest success rate (93.2\%) and lowest collision rate (1.4\%), while also attaining the largest worst-tail CVaR margin. 
It also shows that the proposed method is faster than AIG-MPC and ID-SMPC in both scenarios, but slower than D-MPPI. Relative to RA-MPPI, it is faster in lane merging and slower at intersections. These results indicate that the improvements in success rate and belief clarity are achieved without consistently incurring the highest computational cost.
These results suggest that the proposed framework provides a favorable balance of safety, information acquisition, and task efficiency, although it does not minimize completion time or control effort.

\begin{figure*}
    \centering
\includegraphics[width=0.85\linewidth]{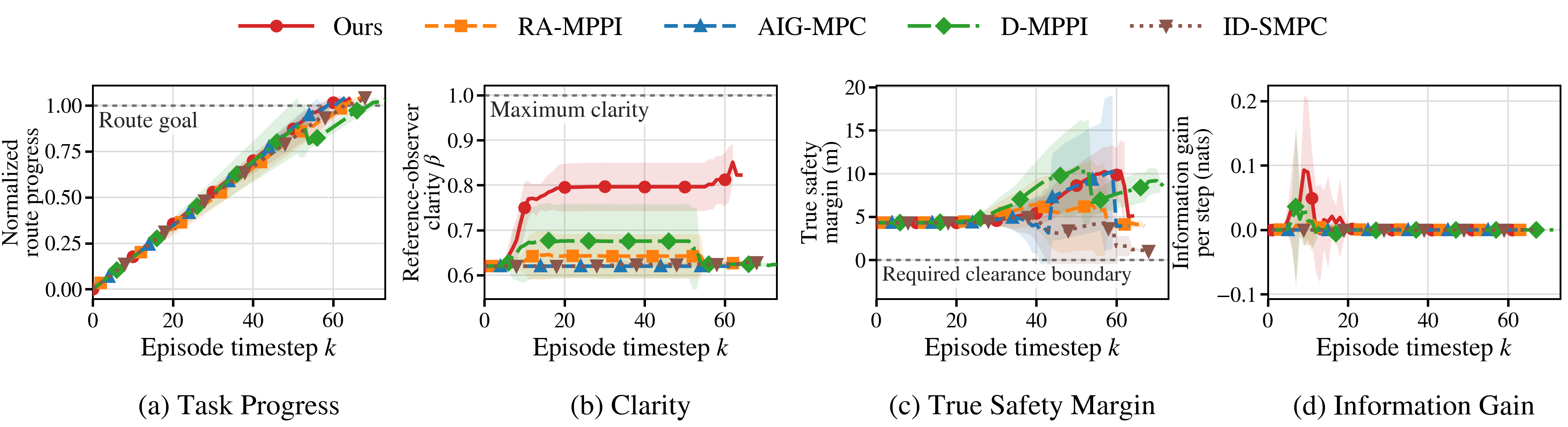}
    \caption{\footnotesize Behavioral comparison across lane-merge runs (mean $\pm$ std).
(a) \textbf{Task progress}: all methods achieve comparable route progress.
(b) \textbf{Clarity}: the proposed method reaches higher reference-observer clarity than the baselines.
(c) \textbf{True safety margin}: mean margins are initially similar but diverge near the merge region; the proposed method maintains a positive mean margin, while several baselines exhibit greater variability or reduced clearance.
(d) \textbf{Information gain}: the proposed method exhibits a pronounced early peak followed by diminishing information gain as clarity increases, consistent with an early emphasis on information acquisition followed by task execution.}
    \label{fig:placeholder}
\end{figure*}

\textbf{Fig.~3} provides further insight into the behavioral differences between planners across lane-merge runs. All methods achieve comparable route progress, while the proposed method reaches higher reference-observer clarity early in the interaction and maintains this advantage throughout the episode. This behavior is consistent with the information objective in the MPC formulation, which encourages actions that improve observation quality. The mean true safety margins are initially similar but diverge near the merge region. The proposed method maintains a positive mean margin, while several baselines exhibit greater variability or reduced clearance.

The information gain per step exhibits a pronounced early peak for the proposed method and subsequently decreases as clarity increases. This pattern is consistent with diminishing returns from information acquisition: as uncertainty is reduced, additional probing provides less benefit, and the controller places greater emphasis on task execution. Together with the comparable route progress, these observations support the benefit of acquiring information early while maintaining task advancement. \textbf{Overall}, these results support coupling belief-space reasoning with clarity and adaptive risk constraints within a unified MPC formulation to balance information acquisition, safety, and efficiency.\vspace{-.2cm}

\subsection{Ablation Studies}
\vspace{-.1cm}

For the ablation study, we analyze how clarity-driven information acquisition and clarity-adaptive CVaR risk regulation influence performance when removed or simplified.
Three reduced variants of the proposed controller are considered. The first is a certainty-equivalent (CE) controller, which ignores uncertainty and treats the mean state estimate as the true system state during planning. This variant corresponds to classical deterministic MPC. The second variant uses fixed CVaR safety margins, enforcing risk constraints independent of belief confidence. This controller captures risk-aware planning without adaptive safety regulation. The third variant enables active information-seeking behavior but removes clarity-adaptive safety constraints, allowing us to examine the benefit of probing without adaptive risk regulation.

Experiments are conducted across a structured set of conditions designed to stress different aspects of the framework. Epistemic uncertainty in the human characteristic is varied across five levels with standard deviation $\sigma_{\theta} \in \{0.3, 0.6, 1.0, 1.5, 2.0\}$. Partial observability is introduced through three occlusion levels $\{0.0, 0.3, 0.6\}$ that scale observation noise in the measurement model, with higher levels representing degraded sensing reliability. Two CVaR confidence levels $\alpha \in \{0.05, 0.1\}$ are evaluated to study sensitivity to tail-risk specification, and four human aggressiveness parameters $\{-0.5, -0.2, 0.2, 0.5\}$ represent cooperative to aggressive driving behavior. Across both lane merging and intersection scenarios, this configuration results in a total of 2,880 simulation runs.

\begin{table*}[t]
\centering
\footnotesize
\caption{\footnotesize Ablation study across all scenarios (2,880 runs). The proposed method achieves the highest success rate (92.9\%) and mean final clarity ($0.85 \pm 0.06$). Fixed CVaR attains the lowest collision rate (0.5\%) but the longest average time ($7.92 \pm 2.31$~s). Certainty-Eq is fastest ($5.01 \pm 1.88$~s) but has the highest collision rate (5.6\%) and negative mean CVaR safety slack, indicating repeated safety margin violations.}
\label{tab:ablation}
\begin{tabular}{lccc>{\columncolor{oursblue}}c}
\toprule
\textbf{Metric} & \textbf{Certainty-Eq} & \textbf{Fixed CVaR} & \textbf{Info-Seeking} & \textbf{Ours} \\
\midrule
Success Rate & $78.4\%$ & $86.7\%$ & $88.2\%$ & $\mathbf{92.9\%}$ \\
Collision Rate & $5.6\%$ & $\mathbf{0.5\%}$ & $2.1\%$ & $0.9\%$ \\
Average Time (s) & $\mathbf{5.01 \pm 1.88}$ & $7.92 \pm 2.31$ & $6.21 \pm 2.04$ & $5.63 \pm 1.76$ \\
Mean Final Clarity $\beta \uparrow$ & $0.61 \pm 0.09$ & $0.58 \pm 0.05$ & $0.72 \pm 0.14$ & $\mathbf{0.85 \pm 0.06}$ \\
Mean CVaR Safety Slack $\uparrow$ & $-0.12 \pm 0.21$ & $\mathbf{0.41 \pm 0.10}$ & $0.18 \pm 0.19$ & $0.32 \pm 0.12$ \\
\bottomrule
\end{tabular}
\end{table*}

\textbf{Table~II} summarizes aggregate performance across all configurations. The certainty-equivalent controller exhibits the lowest success rate and highest collision rate, together with negative mean CVaR safety slack, indicating limitations in planning based solely on the mean estimate. The fixed-margin CVaR controller achieves the lowest collision rate (0.5\%) but produces the longest average completion time (7.92~s), consistent with conservative safety regulation. The information-seeking variant improves final clarity and success rate relative to both variants, supporting the benefit of active probing. The proposed method achieves the highest success rate (92.9\%) and mean final clarity (0.85), while reducing average completion time to 5.63~s and maintaining a low collision rate (0.9\%). These results suggest that combining information acquisition with adaptive risk regulation provides a favorable safety--efficiency balance.

\begin{figure*}
    \centering
    \includegraphics[width=0.85\linewidth]{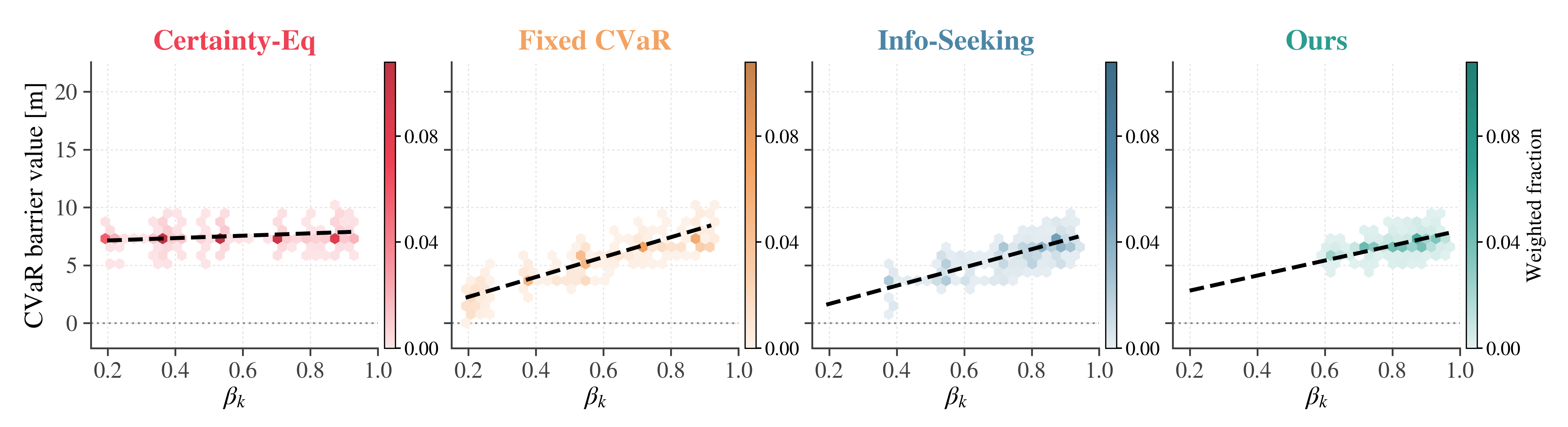}
    \caption{\footnotesize CVaR barrier value versus clarity $\beta_k$ for each ablation variant, across all simulation runs. The proposed method concentrates in the high-clarity region while maintaining positive barrier values, illustrating the combination of belief refinement and safety preservation.}    \label{fig:placeholder}
\end{figure*}

\textbf{Fig.~4} provides further insight into the relationship between clarity and CVaR barrier values. The proposed method concentrates in the high-clarity region while maintaining positive barrier values, consistent with belief refinement alongside safety preservation. Fixed CVaR and Info-Seeking also exhibit positive trends, whereas Certainty-Eq shows comparatively little variation with clarity. Thus, the distinguishing feature of the proposed method is its concentration at high clarity with positive barrier values, rather than a positive correlation alone.
\textbf{Overall}, the ablation results support the complementary roles of clarity-driven probing and clarity-adaptive CVaR safety regulation. Combining these mechanisms within the belief-space MPC formulation enables more reliable interaction while limiting the efficiency cost of conservative planning.
\vspace{-.2cm}
\section{Conclusions}
\label{sec:conclusions} \vspace{-.1cm}
Our work demonstrates the benefit of co-designing safety regulation and information acquisition around a shared representation of belief confidence. To this end, we developed a belief-space MPC framework in which clarity, derived from the Kalman--Riccati covariance recursion, drives probing that resolves the human characteristic and modulates CVaR barrier conservatism. This coupling enables a transition from uncertainty reduction to task-oriented execution as confidence increases, without explicit probing thresholds or mode switching. Across lane merging and unsignalized intersection simulations, the proposed method achieves the highest success rates and lowest collision rates among the evaluated baselines, together with higher belief clarity and competitive completion times.

The current evaluation focuses on two-agent simulations and assumes that the human characteristic remains locally stable during each interaction. Future work will study multi-agent and time-varying behaviors, improve computational efficiency for real-time control, and validate the framework in more diverse traffic conditions. \vspace{.2cm}

{\small
	\bibliographystyle{ieeetr}\footnotesize\vspace{-.4cm}
	\bibliography{ref}

@article{sadigh2018planning,
  title={Planning for cars that coordinate with people: leveraging effects on human actions for planning and active information gathering over human internal state},
  author={Sadigh, Dorsa and Landolfi, Nick and Sastry, Shankar S and Seshia, Sanjit A and Dragan, Anca D},
  journal={Autonomous Robots},
  volume={42},
  number={7},
  pages={1405--1426},
  year={2018},
  publisher={Springer}
}

@article{zhang2024intention,
  title={Intention-aware control based on belief-space specifications and stochastic expansion},
  author={Zhang, Zengjie and Sun, Zhiyong and Haesaert, Sofie},
  journal={IEEE Transactions on Intelligent Vehicles},
  year={2024},
  publisher={IEEE}
}

@inproceedings{zheng2022belief,
  title={Belief space planning: A covariance steering approach},
  author={Zheng, Dongliang and Ridderhof, Jack and Tsiotras, Panagiotis and Agha-mohammadi, Ali-akbar},
  booktitle={2022 International Conference on Robotics and Automation (ICRA)},
  pages={11051--11057},
  year={2022},
  organization={IEEE}
}

@article{chakravarty2026multimodal,
  title={Multimodal Belief-Space Covariance Steering with Active Probing and Influence for Interactive Driving},
  author={Chakravarty, Devodita and Dolan, John and Lyu, Yiwei},
  journal={arXiv preprint arXiv:2602.14540},
  year={2026}
}

@article{cleac2019algames,
  title={ALGAMES: A fast solver for constrained dynamic games},
  author={Cleac'h, Simon Le and Schwager, Mac and Manchester, Zachary},
  journal={arXiv preprint arXiv:1910.09713},
  year={2019}
}

@inproceedings{trautman2010unfreezing,
  title={Unfreezing the robot: Navigation in dense, interacting crowds},
  author={Trautman, Peter and Krause, Andreas},
  booktitle={2010 IEEE/RSJ International Conference on Intelligent Robots and Systems},
  pages={797--803},
  year={2010},
  organization={IEEE}
}

@inproceedings{jasour2025risk,
  title={Risk-Aware Integrated Task and Motion Planning for Versatile Snake Robots Under Localization Failures},
  author={Jasour, Ashkan and Daddi, Guglielmo and Endo, Masafumi and Vaquero, Tiago S and Paton, Michael and Strub, Marlin P and Corpino, Sabrina and Ingham, Michel and Ono, Masahiro and Thakker, Rohan},
  booktitle={2025 IEEE International Conference on Robotics and Automation (ICRA)},
  pages={8880--8886},
  year={2025},
  organization={IEEE}
}

@inproceedings{zhang2023human,
  title={The human gaze helps robots run bravely and efficiently in crowds},
  author={Zhang, Qianyi and Hu, Zhengxi and Song, Yinuo and Pei, Jiayi and Liu, Jingtai},
  booktitle={2023 IEEE international conference on robotics and automation (ICRA)},
  pages={7540--7546},
  year={2023},
  organization={IEEE}
}

@article{yin2025occlusion,
  title={Occlusion-aware interactive end-to-end autonomous driving for right-of-way conflicts},
  author={Yin, Jialun and Zhao, Kun and Ma, Xiaohan and Yan, Siping and Li, Haoran and Yang, Junru and Chen, Yin},
  journal={Machines},
  volume={13},
  number={10},
  pages={965},
  year={2025},
  publisher={MDPI}
}

@inproceedings{knaup2024active,
  title={Active learning with dual model predictive path-integral control for interaction-aware autonomous highway on-ramp merging},
  author={Knaup, Jacob and D’sa, Jovin and Chalaki, Behdad and Naes, Tyler and Mahjoub, Hossein Nourkhiz and Moradi-Pari, Ehsan and Tsiotras, Panagiotis},
  booktitle={2024 IEEE International Conference on Robotics and Automation (ICRA)},
  pages={14191--14197},
  year={2024},
  organization={IEEE}
}

@inproceedings{hu2022active,
  title={Active uncertainty reduction for human-robot interaction: An implicit dual control approach},
  author={Hu, Haimin and Fisac, Jaime F},
  booktitle={International Workshop on the Algorithmic Foundations of Robotics},
  pages={385--401},
  year={2022},
  organization={Springer}
}

@inproceedings{gadginmath2025active,
  title={Active probing with multimodal predictions for motion planning},
  author={Gadginmath, Darshan and Nawaz, Farhad and Sung, Minjun and Tariq, Faizan M and Bae, Sangjae and Isele, David and Pasqualetti, Fabio and D’sa, Jovin},
  booktitle={2025 IEEE/RSJ International Conference on Intelligent Robots and Systems (IROS)},
  pages={3929--3936},
  year={2025},
  organization={IEEE}
}

@misc{rezazadeh2019sub,
  title={A sub-modular receding horizon solution for mobile multi-agent persistent monitoring: animated numerical example},
  author={Rezazadeh, N and Kia, SS},
  year={2019}
}

@article{wang2020planning,
  title={Planning with submodular objective functions},
  author={Wang, Ruosong and Zhang, Hanrui and Chaplot, Devendra Singh and Garagi{\'c}, Denis and Salakhutdinov, Ruslan},
  journal={arXiv preprint arXiv:2010.11863},
  year={2020}
}

@article{zhang2025near,
  title={Near-optimal online learning for multi-agent submodular coordination: Tight approximation and communication efficiency},
  author={Zhang, Qixin and Wan, Zongqi and Yang, Yu and Shen, Li and Tao, Dacheng},
  journal={arXiv preprint arXiv:2502.05028},
  year={2025}
}

@article{naveed2025provably,
  title={Provably Safe Stein Variational Clarity-Aware Informative Planning},
  author={Naveed, Kaleb Ben and Sahai, Utkrisht and Girard, Anouck and Panagou, Dimitra},
  journal={arXiv preprint arXiv:2511.09836},
  year={2025}
}

@inproceedings{yin2023risk,
  title={Risk-aware model predictive path integral control using conditional value-at-risk},
  author={Yin, Ji and Zhang, Zhiyuan and Tsiotras, Panagiotis},
  booktitle={2023 IEEE International Conference on Robotics and Automation (ICRA)},
  pages={7937--7943},
  year={2023},
  organization={IEEE}
}

@article{tangri2026value,
  title={Value-at-Risk Constrained Policy Optimization},
  author={Tangri, Rohan and Calliess, Jan-Peter},
  journal={arXiv e-prints},
  pages={arXiv--2601},
  year={2026}
}

@inproceedings{kishida2024risk,
  title={Risk-aware control of discrete-time stochastic systems: Integrating kalman filter and worst-case cvar in control barrier functions},
  author={Kishida, Masako},
  booktitle={2024 IEEE 63rd Conference on Decision and Control (CDC)},
  pages={2019--2024},
  year={2024},
  organization={IEEE}
}

@inproceedings{lee2022risk,
  title={Risk-sensitive MPCS with deep distributional inverse RL for autonomous driving},
  author={Lee, Keuntaek and Isele, David and Theodorou, Evangelos A and Bae, Sangjae},
  booktitle={2022 IEEE/RSJ International Conference on Intelligent Robots and Systems (IROS)},
  pages={7635--7642},
  year={2022},
  organization={IEEE}
}

@inproceedings{wang2025safe,
  title={Safe navigation in uncertain crowded environments using risk adaptive cvar barrier functions},
  author={Wang, Xinyi and Kim, Taekyung and Hoxha, Bardh and Fainekos, Georgios and Panagou, Dimitra},
  booktitle={2025 IEEE/RSJ International Conference on Intelligent Robots and Systems (IROS)},
  pages={7669--7676},
  year={2025},
  organization={IEEE}
}

@inproceedings{vahs2024risk,
  title={Risk-aware control for robots with non-gaussian belief spaces},
  author={Vahs, Matti and Tumova, Jana},
  booktitle={2024 IEEE International Conference on Robotics and Automation (ICRA)},
  pages={11661--11667},
  year={2024},
  organization={IEEE}
}

@article{agrawal2023sensor,
  title={Sensor-based planning and control for robotic systems: Introducing clarity and perceivability},
  author={Agrawal, Devansh R and Panagou, Dimitra},
  journal={IEEE Control Systems Letters},
  volume={7},
  pages={2623--2628},
  year={2023},
  publisher={IEEE}
}

@article{moon2026ia,
  title={Ia-tigris: An incremental and adaptive sampling-based planner for online informative path planning},
  author={Moon, Brady and Suvarna, Nayana and Jong, Andrew and Chatterjee, Satrajit and Yuan, Junbin and Cao, Muqing and Scherer, Sebastian},
  journal={IEEE Transactions on Robotics},
  year={2026},
  publisher={IEEE}
}

@phdthesis{moon2025informative,
  title={Informative Path Planning Toward Autonomous Real-World Applications},
  author={Moon, Brady},
  year={2025},
  school={Carnegie Mellon University}
}
}
\end{document}